\documentclass{article}
\usepackage{spconf,amsmath,graphicx,array}

\usepackage{enumitem}
\setlist{nosep, leftmargin=14pt}

\usepackage{mwe} 

\title{Chest-Diffusion: A Light-Weight Text-to-Image Model for Report-to-CXR Generation}
\name{Peng Huang$^{a,b,*}$, Xue Gao$^{a,b,*}$, Lihong Huang$^{a,b}$, Jing Jiao$^{a,b}$, Xiaokang Li$^{a,b}$,Yuanyuan Wang\textsuperscript{a,b,†}, Yi Guo\textsuperscript{a,b,†}}

\address{$^{a}$ Department of Electronic Engineering, Fudan University, Shanghai 200433, China \\
$^{b}$ Key Laboratory of Medical Imaging Computing and Computer Assisted Intervention of Shanghai, \\  Shanghai 200032, China
} 

\begin{document}
%
\maketitle
\begin{abstract}
Text-to-image generation has important implications for generation of diverse and controllable images. Several attempts have been made to adapt Stable Diffusion (SD) to the medical domain. However, the large distribution difference between medical reports and natural texts, as well as high computational complexity in common stable diffusion limit the authenticity and feasibility of the generated medical images. To solve above problems, we propose a novel light-weight transformer-based diffusion model learning framework, Chest-Diffusion, for report-to-CXR generation. Chest-Diffusion employs a domain-specific text encoder to obtain accurate and expressive text features to guide image generation, improving the authenticity of the generated images. Meanwhile, we introduce a light-weight transformer architecture as the denoising model, reducing the computational complexity of the diffusion model. Experiments demonstrate that our Chest-Diffusion achieves the lowest FID score 24.456, under the computation budget of 118.918 GFLOPs, which is nearly one-third of the computational complexity of SD.
\end{abstract}

\textit{\textbf{Index Terms}}---Chest X-ray, Diffusion model, Text-to-Image generation, CLIP, Vision Transformer

\footnote{* Contributed Equally, † Corresponding Authors.}

\section{Introduction}
Recent years, image generation has been a hot topic in the computer vision area. The emergence of artificial intelligence contributes to the success of image generation in many fields including super-resolution \cite{b1}, style transfer \cite{b2, b3}, text-to-image generation \cite{b4}-\cite{b10} \emph{et.al}, which helps to enhance visual content creation, improve image quality and foster innovation in computer graphics and design. In medical scenes, dominant image generation methods rely on image-to-image generation, which are largely represented by Generative Adversarial Network (GAN) \cite{b11}. GAN generates images efficiently with a generator-discriminator structure, while the diversity and controllability of generated images are limited due to the challenge of specifying detailed visual attributes solely through image-based methods. Therefore, text-to-image generation, which can generate images with specific descriptions, becomes extremely important.

Text-to-image generation has witnessed the great success in the creation of natural images \cite{b8}-\cite{b10}. An emerging and prominent method is the diffusion model \cite{b12}-\cite{b13}, which uses a Markov chain process to denoise images by modeling them with Gaussian distributions \cite{b14}. As one typical diffusion model, Stable Diffusion (SD) trains models in the latent space of powerful pre-trained autoencoders and introduces cross-attention layers for general conditional inputs, achieving state-of-the-art scores for various image generation tasks \cite{b15}.

To date, two studies have been conducted to adapt SD to text-to-image generation in the medical domain, which mainly focus on generating Chest X-Ray (CXR) images from reports. Chambon \emph{et al}. \cite{b4} investigates whether SD could be adapted to the medical domain in a few-shot manner. RoentGen \cite{b5} thoroughly explores the finetuning strategies of SD, which keeps the autoencoder frozen and finetunes the denoising model as well as the text encoder of the pre-trained SD. Compared with general natural language, medical reports employ a domain-specific vocabulary and a more extended expression. Directly finetuning pretrained text encoders from the natural domain to the medical datasets can result in suboptimal generalization, affecting the quality of generate CXRs. Additionally, SD has a large number of parameters and the convolutional structure of SD requires additional modules when dealing with multi-modal data, which increases computational complexity.

Witnessed the success of the Vision Transformer (ViT) in diverse vision tasks, U-ViT \cite{b16}, a simple and general ViT-based architecture, is widely used for natural image generation with diffusion models, which treats all inputs as tokens and employs long skip connections between shallow and deep layers. It successfully improves the performance of diffusion models in class-conditional and text-to-image generation with fewer parameters than SD. However, the distribution disparities between the medical reports and natural-domain texts limit the direct application of U-ViT in medical domain. 
To this end, we propose a novel transformer-based learning framework of diffusion models for report-to-CXR generation in a high-efficiency manner. Our contributions can be summarized as follows.

1. We propose a light-weight transformer-based framework based on the diffusion models, Chest-Diffusion, to generate realistic CXRs from medical reports and achieves the new state-of-the-art results.

2. A domain-specific CLIP, which learns a joint multi-modal embedding space of reports and CXRs, is introduced to learn more representative and precise report embeddings, thus improving the authenticity of generated CXRs.

3. The U-ViT is adapted to overcome the distributional shift between medical reports and natural-domain texts. The modified U-ViT can concurrently manage multi-modal information, including time, medical reports and images, avoiding the introduction of extra modules and reducing the computational complexity.

4. Results prove that our Chest-Diffusion can generate more realistic images with a smaller computational complexity, surpassing existing diffusion-based report-to-CXR generation methods.

\section{Method}
The overview of proposed Chest-Diffusion is illustrated in Fig. 1. Initially, a domain-specific CLIP learns a joint multi-modal embedding space of reports and CXRs by predicting the correct parings of a batch of (report, CXR) with the contrastive objective. Then, in the report-to-CXR generation process, the pretrained CLIP model are frozen and encode reports into informative textual representation, which are fed to the subsequent denoising model. Finally, the transformer-based denoising model treats the corrupted image latent representations, textual guidance, and temporal information as tokens to predict the clean image representations, all within the existing framework without the need for additional modules. Notably, the diffusion process is applied in the latent space of CXRs, which is conducted by the encoder of the pretrained autoencoder. Predicted image embeddings will be sent to the decoder of the autoencoder to generate the final image.

\subsection{Formulation of Diffusion Model}
Diffusion models \cite{b13} gradually inject noise to image, and subsequently reverse this process to generate image from noise. The noise injection process, called the forward process, is formalized as a Markov chain of length T: 
\begin{equation}
q(x_{1:T}|x_{0})=\prod_{1}^{T}q(x_{t}|x_{t-1})
\end{equation} where ${x}_0$ is the input uncorrupted image, $q(x_{t}|x_{t-1})=N(x_{t}|\sqrt{1-\beta_{t}}x_{t-1}, \beta_{t}I)$, with time-varying variance $\beta_{t}\in(0, 1)$, selected according to a predefined variance schedule.

\begin{figure}[!h]
\centerline{\includegraphics[width=\columnwidth]{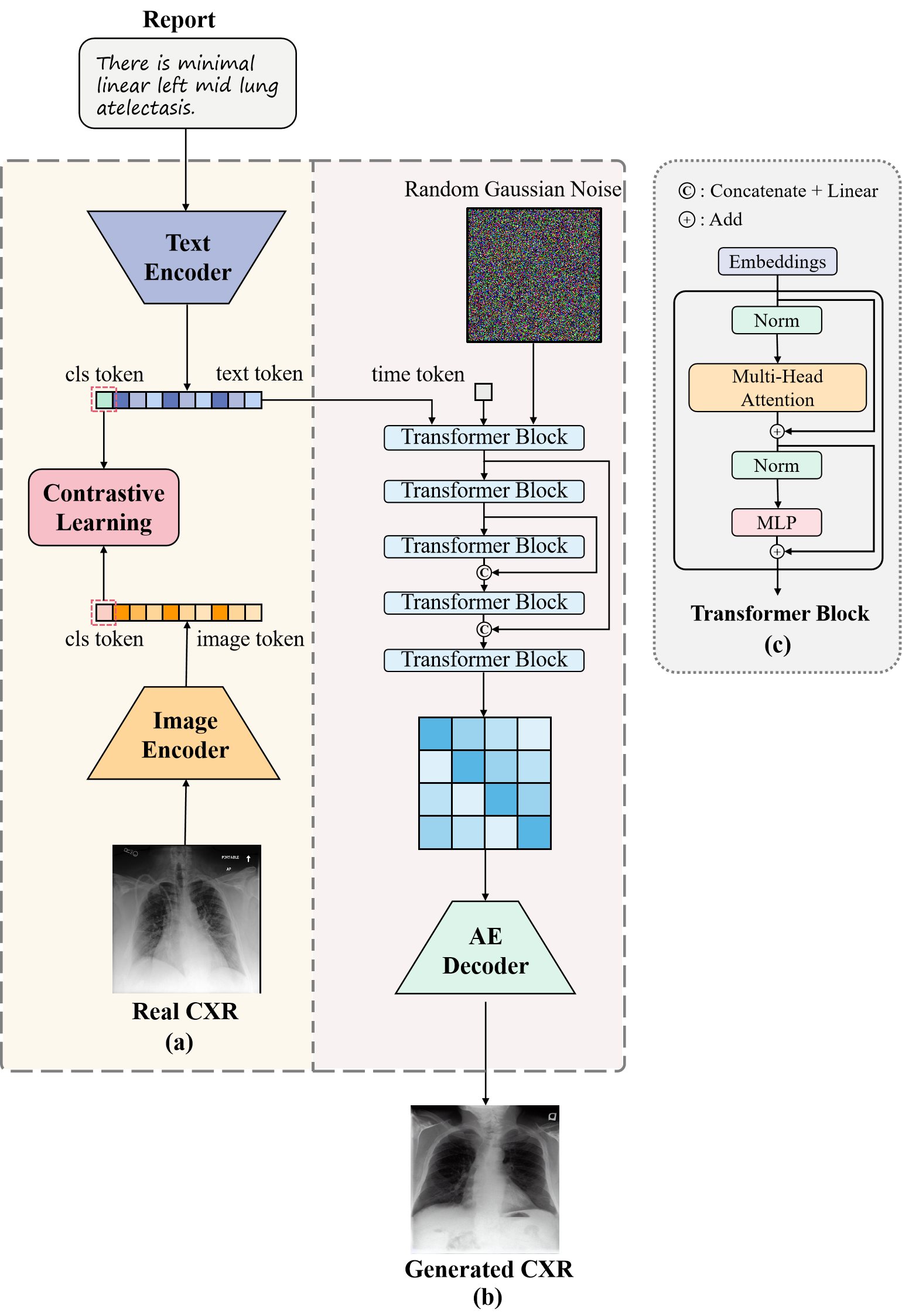}}
\caption{Overview of the proposed Chest-Diffusion. (a) shows the training process of domain-specific CLIP. (b) depicts the report-to-CXR generation process. (c) shows the specific structure of Transformer Block.}
\label{fig1}
\end{figure}

The learning task is to reverse this forward process by deducing $x_{t-1}$ from its corrupted version $x_{t}$ for $t\leq T$. The corresponding objective can be expressed as:
\begin{equation}
\min_{\theta}E_{t,x_{0},c,\epsilon}=\| \epsilon - \epsilon_{\theta}(x_{t},t) \|_{2}^2
\end{equation}
where ${\epsilon}_{\theta}(x_{t}, t)$ denotes a sequence of denoising autoencoders parameterized with $\theta$ sharing equal weights to predict the noise $\epsilon$.

\begin{figure*}
\centerline{\includegraphics[width=1.9\columnwidth]{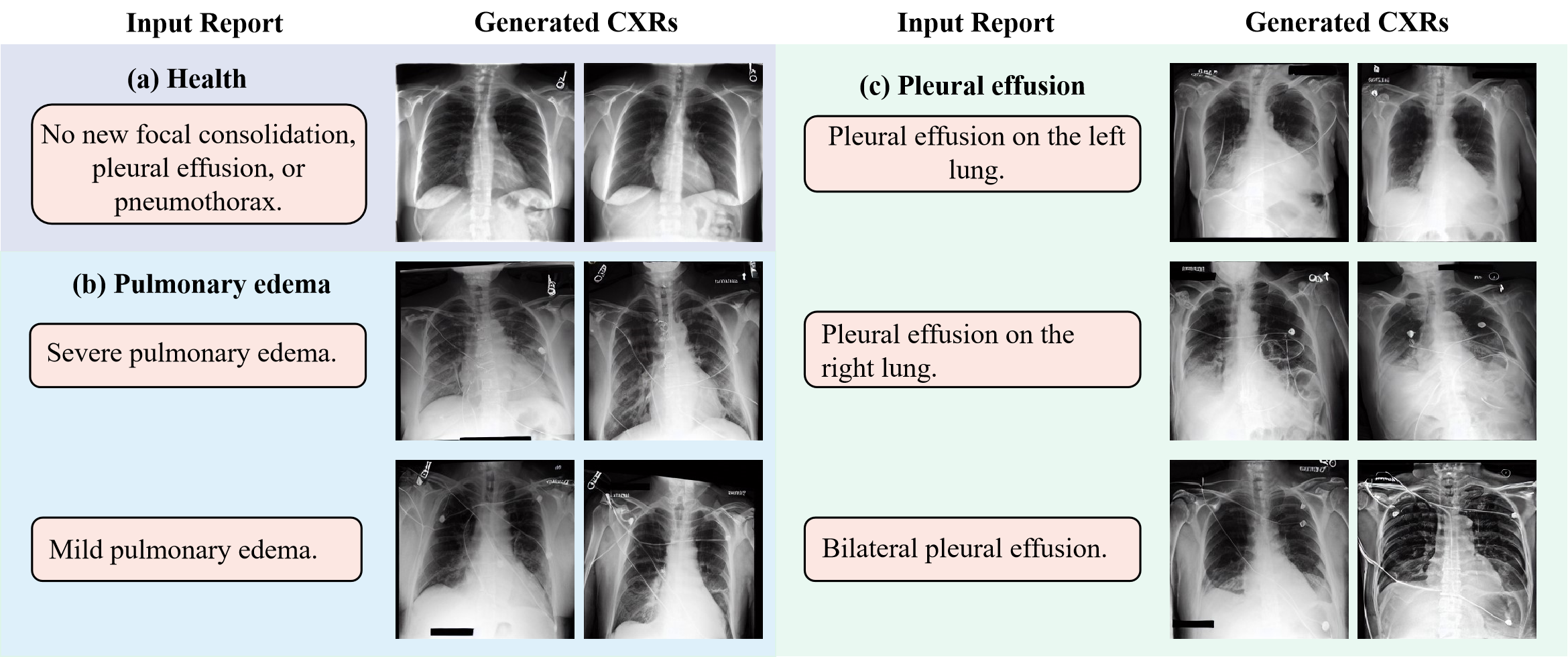}}
\caption{CXRs generated by Chest-Diffusion using different reports as input. (a) CXR of healthy people. (b) CXRs of patients with varying degrees of pulmonary. (c) CXR of patients with pleural effusion in different location.}
\label{fig2}
\end{figure*}

\subsection{Chest-Diffusion}
Chest-Diffusion applies the diffusion process in the latent space of CXRs, which is a light-weight, transformer-based learning framework designed for report-to-CXR generation. Our Chest-Diffusion mainly includes three essential components, the CLIP-based text encoder, the pretrained autoencoder and the transformer-based denoising model, which are detailed follows.

{\bfseries CLIP-based text encoder.} To obtain the informative features of medical reports to facilitate the generation of realistic and diverse CXRs from reports, we employ the domain-specific CLIP model, BiomedCLIP \cite{b17}, as our text encoder. It was trained under 15 million biomedical image-text-pairs (PMC-15M) and displays remarkable proficiency in extracting medical report features. Similar to CLIP \cite{b18}, BiomedCLIP constructs a multi-modal embedding space through concurrent training of an image encoder and a text encoder with the contrastive objective. The contrastive objective maximizes the cosine similarity between the embeddings of correct image-text pairs in a batch while minimizing the similarity between incorrect pairs. To better model the lengthy medical reports, the tokenizer limits are increased to 256, avoiding unnecessary truncation. In addition, to learn more fine-grained text features and establish a better initial connection between reports and CXRs, we further finetune the BiomedCLIP on our dataset with the contrastive objective mentioned above.

{\bfseries Pretrained autoencoder.} In this module, our objective is to construct a representative latent space for input CXRs, where the diffusion process is modeled. It has been demonstrated that the autoencoder within the SD architecture, trained using a combination of perceptual loss and an adversarial objective, can acquire robust latent representations of CXRs and generate realistic images \cite{b4}. Consequently, we leverage the pretrained SD autoencoder component, keeping it frozen, to derive the latent representations of input images. Given an input CXR $x \in R^{H \times W \times 3}$, the encoder encodes $x$ into the latent representation $z \in R^{h \times w \times 8}$. Notably, the encoder usually downsamples the CXR by a factor $s=H / h= W / w$ and $s$ is set as 8 in our experiment. The decoder can reconstruct the CXR from the latent obtained by the encoder.

{\bfseries Transformer-based denoising model.} In order to reduce the parameters and computational complexity of the conditional denoising model, we adapt the transformer-based architecture, U-ViT, for report-to-CXR generation. Hence, the denoising model can take time, noisy image representation and text condition as input tokens simultaneously to reconstruct the original image representation, without introducing additional modules. In our experiments, the U-VIT we adopted contains 8 encoder transformer blocks, 8 decoder transformer blocks and 1 middle transformer block. To adjust U-ViT for processing medical reports with a greater number of tokens, we augment the dimensionality of U-ViT’s positional encoding layer. Position embeddings will be added to the time embeddings, noisy image representation and the report embeddings to preserve position information. Subsequently, the transformer blocks model the intermodal correlations between multi-modal information and deduce the clean image representation. The training objective of this module can be formulated by minimizing
\begin{equation}
\min_{\theta}E_{t,z_{0},c,\epsilon}=\| \epsilon - \epsilon_{\theta}(z_{t},t) \|_{2}^2
\end{equation}

\begin{table*}[t]
\centering
\caption{Comparison with existing studies based on the computational complexity and the realism of generated CXRs.}
\label{table}
\setlength{\tabcolsep}{5pt}
\begin{tabular*}{\textwidth}
{@{\extracolsep{\fill}}>
{\centering\arraybackslash}p{100pt}> {\centering\arraybackslash}p{20pt}> {\centering\arraybackslash}p{20pt}> {\centering\arraybackslash}p{20pt}> {\centering\arraybackslash}p{20pt}}
\hline
\centering
 & Parameters (M) & Inference time(s) & Complexity (GFLOPS) & FID \\
\hline
RoentGen & 865.785 & 5.185 & 345.229 & 56.499 \\
LLM-CXR & - & 9.780 & - & 29.270\\
Chest-Diffusion & \textbf{58.175} & \textbf{1.828} & \textbf{118.918} & \textbf{24.456}\\
\hline
\end{tabular*}
\label{tab1}
\end{table*}

\begin{table*}[h]
\centering
\caption{Comparison with existing studies based on the vision-language alignment between the input report and generated CXR. Atel.: Atelectasis, Cnsl.: Consolidation, PTX.: Pneumothorax, Edma.: Edema, Eff.: Pleural Effusion, Pneum.: Pneumonia, Cmgl.: Cardiomegaly, Les.: Lung Lesion, Fra.: Fracture, Opac.: Lung Opacity, Encar.: Enlarged Cardiomediastinum, Avg: Average.}
\label{table}
\setlength{\tabcolsep}{5pt}
\begin{tabular*}{\textwidth}{@{\extracolsep{\fill}}>{\centering\arraybackslash}p{100pt}> {\centering\arraybackslash}p{20pt}> {\centering\arraybackslash}p{20pt}> {\centering\arraybackslash}p{20pt}> {\centering\arraybackslash}p{20pt}> {\centering\arraybackslash}p{20pt}> {\centering\arraybackslash}p{20pt}> {\centering\arraybackslash}p{20pt}> {\centering\arraybackslash}p{20pt}> {\centering\arraybackslash}p{20pt}> {\centering\arraybackslash}p{20pt}> {\centering\arraybackslash}p{20pt}> {\centering\arraybackslash}p{20pt}}
\hline
  & Atel. & Cnsl. & PTX. & Edma. & Eff. & Pneum. & Cmgl. & Les. & Frac. & Opac. & Encar. & Avg.\\
\hline
RoentGen & 0.590 & 0.505 & 0.523 & 0.602 & 0.599 & 0.518 & 0.648	& 0.474	& \textbf{0.574} & 0.554 & 0.530 & 0.556\\
LLM-CXR & 0.662 & 0.602 & 0.722 & 0.754 & 0.781 & 0.546 &	0.661 &	0.477 &	0.482 &	0.640 &	0.581 &	0.628\\
Chest-Diffusion & \textbf{0.696} & \textbf{0.627} & \textbf{0.693} & \textbf{0.787} & \textbf{0.845} & \textbf{0.565} & \textbf{0.730} & \textbf{0.527} & 0.526 & \textbf{0.648} & \textbf{0.597} & \textbf{0.658}\\
\hline
\end{tabular*}
\label{tab3}
\end{table*}

\section{Experiments}
\subsection{Dataset and evaluation}

To develop Chest Diffusion, we employed the publicly available dataset, MIMIC-CXR \cite{b19}, provided by the Beth Israel Deaconess Medical Center. The dataset includes 377,110 chest X-ray images and 227,835 reports. Each image is labeled with one or multiple classes of 14 diagnostic labels. The official data splits (i.e., 70\%/10\%/20\% for train/validation/test set) are adopted. Only anterior-posterior (AP)/posterior-anterior (PA) scans are considered and duplicates are removed in our analysis. Hence, our dataset consisted of 162,914 training samples, 1,286 validation samples, and 2,461 testing samples. All images are resized into $256 \times 256$.
We evaluate the performance of Chest-Diffusion from three perspectives: computational complexity, quality of generated images, and alignment between vision and language. Floating Point Operations (FLOPs) are in wide use to represent the computational complexity of different methods. In our experiment, we set the batch size as 4 to calculate FLOPS and further make a statistic of parameters and inference time of different models to extensively evaluate the computational complexity. Fewer parameters, shorter inference time, and lower FLOPS collectively indicate lower computational complexity. To measure the quality of generated images, we utilize Fréchet Inception Distance (FID) \cite{b20}, which quantifies the distribution discrepancy between generated data and real data and evaluates the authenticity of generated data. A lower FID score indicates the better authenticity of generated data. Furthermore, for a comprehensive evaluation, we measure vision-language alignment by calculating the AUROC against the original CheXpert \cite{b21} labels from MIMIC-CXR with a pretrained CXR disease classifier network, densnet121-res224-all \cite{b22}.  Higher AUROC score means better vision-language alignment between the input report and generated CXR.

\subsection{Comparison with the State-of-the-Art Method}

We compared our Chest-Diffusion with the current solitary report-to-CXR diffusion model RoentGen, which investigated the fine-tuning strategies of SD for the report-to-CXR generation task. We also compare it with the current SOTA autoregressive report-to-CXR model LLM-CXR \cite{b7}.
From {\bfseries TABLE I}, it is evident that Chest-Diffusion surpasses RoentGen and LLM-CXR in generating more realistic CXRs with lower computational complexity. Specifically, Chest-Diffusion achieves the lowest FID score of 24.456, which is nearly half of that of RoentGen, strongly demonstrating the realism of our generated images.  Notably, the optimization of Chest-Diffusion in parameters and computational complexity significantly enhances the inference speed of the diffusion model. The generation speed for each image is about 2.8 times and 5.4 times faster than that of RoentGen and LLM-CXR.  
As shown in {\bfseries TABLE II}, Chest-Diffusion outperforms RoentGen and LLM-CXR in terms of the average AUROC metric, with a margin of 0.102 and 0.030. The AUROC scores of the generated images for nine diseases are consistently higher than those of RoentGen and LLM-CXR, indicating better alignment between the generated images and the input text. What’s more, as illustrated in Fig. 2, Chest-Diffusion is capable of generating realistic CXRs with various lesions of different severity levels and locations.

\section{Conclusion}
This paper proposes a light-weight transformer-based learning framework of diffusion models, Chest-Diffusion, for report-to-CXR generation. Chest-Diffusion relies on the domain-specific CLIP model to learn the informative and accurate text embeddings to guide the report-to-CXR generation. Instead of directing finetuning SD, we modified the U-ViT to make it suitable for processing medical reports which greatly reduces the computational complexity. Results on the public dataset show that Chest-Diffusion achieves the SOTA performance in report-to-CXR generation with less computational complexity. We hope that our Chest-Diffusion can be beneficial for the primary medical education.

\section{COMPLIANCE WITH ETHICAL STANDARDS}
This research study was conducted retrospectively using human subject data made available in open access by MIMIC-CXR. Ethical approval was not required as confirmed by the license attached with the open access data.

\section{ACKNOWLEDGMENTS}
This work was supported by the National Natural Science Foundation of China (Grant 81830058), the Science and Technology Commission of Shanghai Municipality (Grant 22ZR1404800)


\end{document}